\documentclass[lettersize,journal]{IEEEtran}
\usepackage{amsmath,amsfonts}
\usepackage{algorithmic}
\usepackage{algorithm}
\usepackage{array}
\usepackage{textcomp}
\usepackage{stfloats}
\usepackage{url}
\usepackage{verbatim}
\usepackage{graphicx}
\usepackage{cite}
\usepackage{subcaption}
\usepackage{placeins}
\usepackage{multirow}

\begin{document}

\title{Spatial Attention Noise Masking for Causally Sufficient Interpretability}

\author{Benjamin Formby, Kuang-Ching Wang, and D Hudson Smith
\thanks{This work was supported in part by NSF Project (1934966, 2029260, 2027208, 2330891) and NIH Project P20 GM121342.}
\thanks{Benjamin Formby is with the Electrical and Computer Engineering department at Clemson University, Clemson, SC (email: bformby@clemson.edu}
\thanks{Kuang-Ching Wang is with the School of Computing at Binghamton University, Binghamton, NY (email: kc.wang@binghamton.edu}
\thanks{D Hudson Smith is with the Department of Mathematical and Statistical Sciences at Clemson University, Clemson, SC (email: dane2@clemson.edu} }

\markboth{IEEE Transactions on Image Processing}%
{Shell \MakeLowercase{\textit{et al.}}: A Sample Article Using IEEEtran.cls for IEEE Journals}


\maketitle

\begin{abstract}
We present a novel causal approach to interpretability for computer vision models that dynamically masks the input image prior to classification. The interpretability of deep learning predictions is critical in high-stakes fields such as medical imaging, security, and autonomous driving. Most interpretability methods are applied passively to already trained models, which typically result in correlational rather than causal explanations. Existing causal interpretability methods are limited to post hoc analysis, weakening the causal claims. Additionally, existing active methods generally lack explanations that explicitly assign responsibility to input features. This work proposes a spatial attention noise masking framework that provides causal explanations about the features sufficient for the prediction. The proposed framework consists of: 1) a UNet-style mask generator, and 2) a Resnet18 encoder and linear classifier that classifies both masked and unmasked versions of an input image. The generated masks are regularized to be sparse and spatially smooth, while masked image embeddings are constrained to remain consistent with embeddings from the corresponding unmasked images. The resulting masks can be interpreted as feature attribution maps that are competitive with related interpretability methods while additionally providing strong causal explanations of model predictions. Quantitative evaluations demonstrate mask faithfulness, near-baseline classification performance across five classification tasks despite substantial masking of image information, and robustness to distribution shifts such as background swapping and natural adversarial examples. Qualitative comparisons further demonstrate mask behavior and competitive interpretability relative to state-of-the-art feature attribution methods.
\end{abstract}


\section{Introduction} 
\IEEEPARstart{D}{eep} learning has achieved remarkable performance across a wide range of computer vision tasks, including image classification. However, as these models are increasingly deployed in real-world applications, the demand for interpretable decision-making has grown alongside their predictive capabilities \cite{zhang_survey_2021}. In domains such as medical imaging, it is essential to understand whether a tumor is classified as malignant based on clinically meaningful evidence or on spurious artifacts \cite{salahuddin_transparency_2022}. Similarly, in autonomous driving, cybersecurity, and other safety-critical applications, interpretable models enable analysis of failure cases, leading to more reliable decision-making \cite{kuznietsov_explainable_2024, dwivedi_explainable_2023, guo_lemna_2018}. Beyond improving trust and accountability, interpretability can also enable scientific discovery by revealing meaningful patterns and representations learned by deep neural networks \cite{xu_interpretability_2025}. In these high-stakes settings, explanations must not only be interpretable but also faithful to the model's underlying decision-making process \cite{jacovi_towards_2020}.

There have been numerous efforts to explain the predictions and behavior of deep learning models through interpretability methods \cite{ribeiro_why_2016, selvaraju_grad-cam_2017, sundararajan_axiomatic_2017, springenberg_striving_2015}. Most existing approaches for image classification are post hoc, meaning they are applied passively after training and do not influence the optimization process or the representations learned by the model. Consequently, these methods primarily provide correlational explanations by estimating the importance of input features without directly establishing whether those features were responsible for the prediction. Correlational explanations answer questions such as, ``Which features are associated with this prediction?'' In contrast, causal explanations seek to answer questions such as, ``Which features were responsible for this prediction?'' Such explanations provide stronger evidence that the highlighted features were determinant in the model's decision-making process \cite{xu_interpretability_2025}.

To address the limitations of correlational explanations, several methods have explored causal and active forms of interpretability. Many of these causal approaches examine changes in output in response to changes in the input \cite{zeiler_visualizing_2014}; however, these methods are still typically post hoc and rely on counterfactual explanations that merely answer questions like ``How does the prediction change if we remove these features?'' While these methods can provide stronger evidence of feature importance than gradient-based attribution techniques, they are often computationally expensive and remain post hoc, offering explanations only after training has been completed. Active interpretability methods instead incorporate explanation objectives directly into the learning process through attention mechanisms or regularization strategies, encouraging models to rely on interpretable evidence during optimization \cite{vaswani_attention_2023, dosovitskiy_image_2021, plumb_regularizing_2020, weinberger_learning_2020}. However, while these approaches can generate higher-quality, more interpretable explanations, they cannot explicitly assign responsibility to highlighted features, limiting their capacity to provide causal explanations.

\begin{figure*}[t]
\centerline{\includegraphics[width=2\columnwidth]{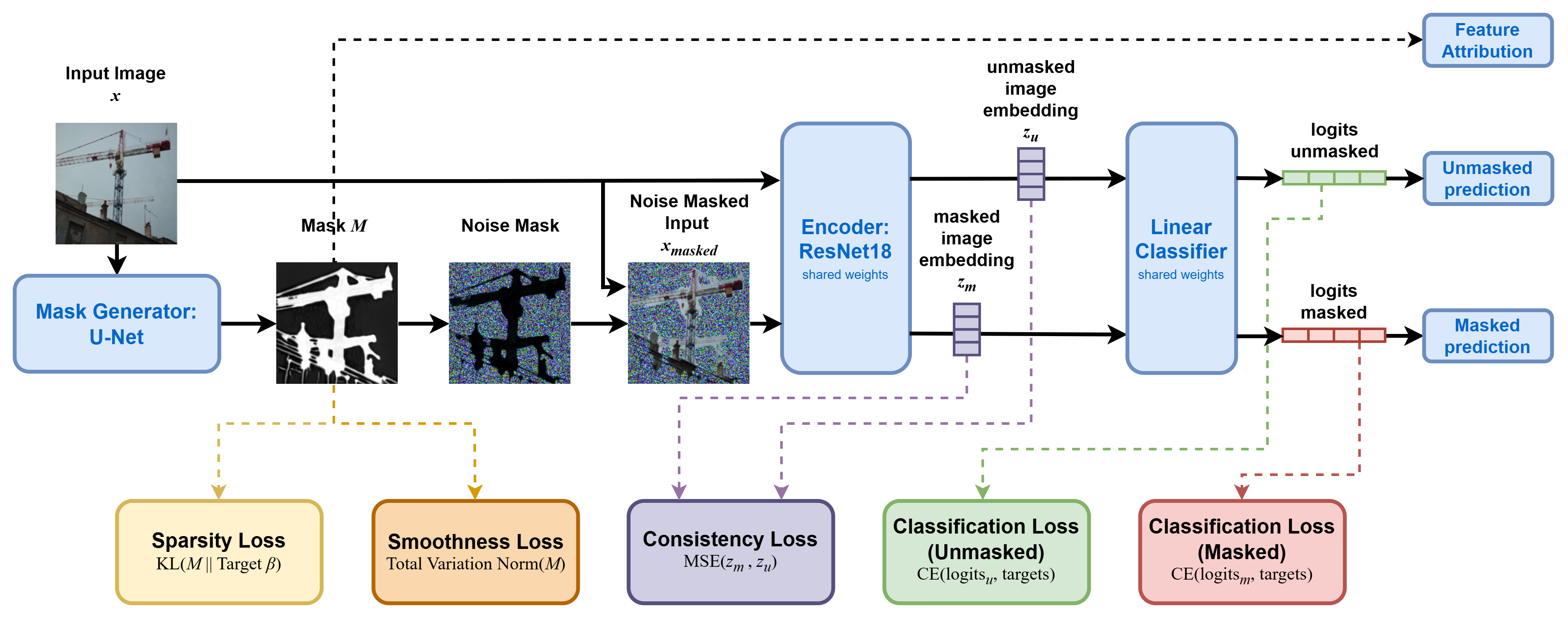}}
\caption{This work proposes a spatial attention noise mask that provides causal explanations of model predictions. We achieve this by generating a mask with image-specific noise added to the masked regions and then applying it to the input image before classification. The classifier is trained on masked and unmasked versions of an input image. The generated masks are regularized to be sparse and spatially smooth, while masked image embeddings are constrained to remain consistent with embeddings from the corresponding unmasked images.}
\label{fig:abstract}
\end{figure*}

This work introduces a spatial attention noise-masking framework for causal interpretability, where a learned mask is applied to the input prior to classification to explicitly identify the image regions that are sufficient for a prediction. By learning to selectively preserve informative regions while suppressing irrelevant ones, the framework attributes responsibility for the output to specific input features. As shown in Figure \ref{fig:abstract}, our framework consists of a mask generator and classifier that are jointly trained with a novel regularization strategy. The mask generator is constrained to achieve the desired sparsity of masked pixels via a Kullback-Leibler (KL) divergence between beta distributions, and constrained by a total variation loss for spatial smoothness. Additionally, the classifier is trained on both masked and unmasked images to maintain baseline classification performance, using a consistency regularization that encourages masked-image embeddings to remain consistent with unmasked-image embeddings.

We summarize the contributions of this work in the following three points:
\begin{itemize}
    \item We introduce Spatial Attention Noise Masking (SANM), a novel approach to interpretability for computer vision models that embeds the feature attribution map directly into the information-processing architecture, thereby providing causally sufficient visual explanations. 
    \item A joint training and regularization strategy that enforces sparse and spatially coherent masks while preserving near-baseline classification performance under substantial input masking.
    \item An extensive evaluation of SANM that demonstrates competitive visual explanations, mask faithfulness, near-baseline classification performance and improved robustness to distribution shifts involving background variations.
\end{itemize}

\section{Related Work}

\subsection{Passive and Active Interpretability}
Interpretability methods for deep learning can be divided into two categories: passive and active \cite{zhang_survey_2021}. Passive approaches are post hoc analyses performed on already trained models \cite{selvaraju_grad-cam_2017, sundararajan_axiomatic_2017, springenberg_striving_2015, zeiler_visualizing_2014}, whereas active approaches modify the model to improve interpretability through a change in architecture or optimization \cite{dosovitskiy_image_2021, khandelwal_attentionrnn_2019, weinberger_learning_2020, plumb_regularizing_2020}. 

Many passive approaches are gradient-based feature attribution methods that generate heatmaps assigning importance to pixels based on the gradient behavior under different conditions. Guided Backpropagation modifies the standard backpropagation process by propagating only positive gradients through neurons with positive activations, producing fine-grained attribution maps \cite{selvaraju_grad-cam_2017}. To address the noise and instability often associated with raw gradients, Integrated Gradients accumulates gradients along a path from a neutral baseline image to the input image, providing a more robust estimate of feature importance \cite{sundararajan_axiomatic_2017}. GradCAM generates coarser attribution maps by weighting feature maps from a target layer using the average gradient of the target class with respect to those feature maps \cite{selvaraju_grad-cam_2017}. Guided GradCAM combines the localization capabilities of GradCAM with the fine-grained detail of Guided Backpropagation to produce higher-resolution visual explanations \cite{selvaraju_grad-cam_2017}. While these methods can highlight features associated with a prediction, they remain passive, post hoc techniques that provide correlational explanations rather than causal insight into the model's decision-making process.

Some passive methods attempt to establish stronger causal evidence by using image perturbations to examine changes in the output in response to changes in the input. For example, Occlusion generates heatmaps of feature importance by systematically masking regions of an image in a sliding window and measuring the change in the model's confidence in the target class \cite{zeiler_visualizing_2014}. Although such perturbation-based methods demonstrate a cause-and-effect relationship between input changes and model predictions, they remain limited to post hoc counterfactual analyses based on hypothetical changes at inference rather than during the learning process. 

Active approaches are less common and tend to use interpretability regularization terms. One of these methods leverages prior knowledge of features, or meta-features, to create more meaningful attributions \cite{weinberger_learning_2020}, while another uses interpretability quality metrics to encourage higher-quality and more stable feature attributions \cite{plumb_regularizing_2020}. However, neither of these active approaches has been extensively evaluated in a spatial context, and both still lack a direct mechanism for assigning responsibility to features that determine the model's decision.

\subsection{Attention-based Interpretability}
Attention mechanisms are another family of active interpretability methods that generate spatial explanations during the prediction process. For example, the self-attention mechanism of a vision transformer can be used to visualize spatial explanations of pixel importance \cite{dosovitskiy_image_2021}. Similarly, \cite{khandelwal_attentionrnn_2019} uses convolutional neural network feature maps and a recurrent neural network to produce structured, coherent attention maps. While these methods provide spatial explanations that identify the regions with greater attention, the attention maps alone do not provide causal explanations.

\subsection{Masking-based Interpretability}
Learned masking approaches are most commonly used in image segmentation; however, some works have explored learned masks as a mechanism for interpretability. Similar to perturbation-based attribution methods, \cite{fong_interpretable_2017} optimizes a sparse mask that perturbs selected image regions to identify features that most strongly influence a model's prediction. Building on this idea, \cite{fong_understanding_2019} introduces smooth masks with fixed-area constraints to improve the stability and consistency of perturbation-based explanations. By directly optimizing masks, these methods often produce more reliable and interpretable explanations than traditional perturbation approaches, such as Occlusion. However, they remain post hoc analyses that require optimization for each individual image, making them computationally expensive and limiting their applicability as active interpretability methods.

\subsection{KL Divergence}
Kullback Leibler (KL) divergence is a method of measuring the difference between an estimated probability distribution and a true probability distribution. While popularly used for generative models  \cite{kingma_introduction_2019,kingma_auto-encoding_2022, goodfellow_generative_2014, ho_denoising_2020}, KL divergence has been used as a sparsity regularization in sparse autoencoders \cite{ng2011sparseautoencoder}. In sparse autoencoders, a KL-divergence penalty encourages hidden neurons to activate only rarely by matching their average activation rate to a sparse Bernoulli prior distribution. These works motivated our use of KL divergence as a sparsity regularizer for spatial attention masks. Unlike simpler mask sparsity regularizations such as L1, which can be minimized by simply dimming the mask and thereby weakening interpretability, KL divergence can better encourage the mask to fit a desired sparsity. Furthermore, to the best of our knowledge, we are the first to employ KL divergence with a Beta prior as a mask sparsity regularizer.

\section{Method}
\label{sec:method}
As shown in Figure \ref{fig:abstract}, the overall framework begins with mask generation of the input image. Masking consists of computing image-matched noise, adding it to the masked regions to produce a noise mask, and applying the noise mask to the image. The noise-masked and unmasked input images are passed through a shared classification network in two forward passes to obtain predictions for both. The masking network and the classification network are trained simultaneously with classification losses, a mask sparsity loss, a mask smoothness loss, and an image embedding consistency loss. The details of the mask generation, classification network, and regularization are described in the following sections.

\subsection{Mask Generation}
\textbf{Architecture} The mask generator used a UNet architecture similar to the architecture described in \cite{ronneberger_u-net_2015}. Given an input image, the UNet produces a single-channel mask using a symmetrical encoder-decoder architecture with skip connections. The encoder captures image context at multiple scales, and the decoder provides localized spatial masks that preserve fine-grained structural details via skip connections.

\textbf{Noise Mask}
Rather than occluding the masked regions of an image, which creates a significant shift in the data distribution of the input images, we add image-specific noise to the complementary regions of the mask before applying the mask to the input image. This computation is given by Equation \ref{eq:mask} where $M$ is the mask, $N_\text{matched}$ is the image-matched noise, $x$ is the input image, and $x_\text{masked}$ is the resulting noise-masked input. 

\begin{equation}\label{eq:mask}
x_\text{masked} =
M \odot x + (1-M) \odot N_\text{matched}
\end{equation}

The image-matched noise is calculated by scaling Gaussian noise by the input image's standard deviation and shifting it by the image mean. The image-matched noise, denoted by $N_\text{matched}$, is defined by Equation \ref{eq:noise} where $N_\text{Gaussian}$ is the Gaussian noise and $x$ is any given input image.

\begin{equation}\label{eq:noise} 
N_\text{matched} = N_\text{Gaussian} \text{SD}(x) +\text{MEAN}(x)
\end{equation}

\textbf{Mask Regularization}
We introduce two loss terms that are computed directly from the generated masks. The first encourages mask sparsity, and the second encourages mask smoothness.

\subsubsection{Sparsity} The mask sparsity must be explicitly constrained because classification loss alone does not encourage sparse masks; instead, the classifier will exploit as much image information as possible to maximize prediction accuracy, likely yielding solid masks that keep the entire image. Therefore, the desired mask pixel intensities should follow a distribution with probability masses concentrated near 0 and 1, producing a near-binary mask with soft edges, but still preserving important structures. A binary-like mask is desirable for causal interpretability because it clearly identifies the regions sufficient for prediction, while soft edges preserve graded importance information for boundary and transition regions. A beta distribution where alpha and beta parameters are less than 1 fits this desired distribution of mask pixel intensities well, so we implement a sparsity loss term, defined as $L_\text{sparsity}$, which penalizes the KL divergence between an empirically estimated beta distribution of the mask pixel intensities and a target beta distribution. Other sparsity regularizers, such as L1 regularization, can often be satisfied by uniformly dimming the image rather than learning truly selective feature masks, limiting interpretability. In contrast, distribution-based regularization encourages mask values to concentrate near 0 and 1, yielding more selective, semantically meaningful masks that better support causal interpretability. $L_\text{sparsity}$ is defined by Equation \ref{eq:kl}. 
\begin{equation}\label{eq:kl} 
L_\text{sparsity} = \frac{1}{N}\sum_{i=1}^{N}  \text{KL}(\text{Beta}(\alpha_f,\beta_f) \,\|\, \text{Beta}(\alpha_g,\beta_g))_i
\end{equation}
$L_\text{sparsity}$ is computed by the mean KL divergence of a batch during training. In this formulation, $\text{Beta}(\alpha_f,\beta_f)$ denotes the empirical beta distribution fitted to the mask pixel intensities of the $i^{th}$ sample in a batch of size $N$, with parameters $\alpha_f$ and $\beta_f$ estimated using the method of moments. $\text{Beta}(\alpha_g,\beta_g)$ denotes the target beta distribution with target parameters $\alpha_g$ and $\beta_g$. For all of our experiments, we use target parameters $\alpha_g=0.2$ and $\beta_g=0.5$.

\subsubsection{Smoothness} Mask smoothness was also constrained to preserve coherent edges and encourage graded importance information in boundary and transition regions. This constraint was implemented using a total variation loss \cite{tvnorm}, denoted by $L_\text{smoothness}$ and defined by Equation \ref{eq:tv}. Total variation loss penalizes large differences between neighboring pixels, reducing high-frequency noise while preserving sharp structural boundaries within the mask. This approach was chosen over alternative smoothing methods, such as Gaussian blurring, because stronger smoothing operations can overly diffuse mask boundaries and remove important spatial detail necessary for interpretability.
\begin{equation}\label{eq:tv}
\begin{split}
L_\text{smoothness} =\;&
\frac{1}{N_h}
\sum_{i,j}
\left(M_{i+1,j}-M_{i,j}\right)^2 \\
&+
\frac{1}{N_w}
\sum_{i,j}
\left(M_{i,j+1}-M_{i,j}\right)^2
\end{split}
\end{equation}
In this formulation, $M$ is the mask, $N_h$ is the number of vertical differences, $N_w$ is the number of horizontal differences, and $i,j$ are the pixel coordinates.

\subsection{Classification Network}
\textbf{Architecture} For classification, we used a ResNet-18 \cite{he_deep_2016} backbone because of its proven performance in image classification tasks. After generating the mask as described in the previous section and applying it to the input image, both the masked and unmasked images proceed through a forward pass of the classifier, resulting in predictions for the masked and unmasked versions of the input image.

\textbf{Classification Regularization}
We introduced three loss terms that come from the classification network. Two are classification losses, and the third is an embedding consistency loss that encourages consistent image embeddings between the masked and unmasked images.

\subsubsection{Classification Losses}
Since there are two forward passes of the classification model for the masked and unmasked images, there is a classification loss for each prediction denoted by $L_\text{masked}$ and $L_\text{unmasked}$, respectively. These classification losses are calculated by the cross-entropy between the predictions and the target class.

\subsubsection{Embedding Consistency Loss}
The embedding consistency loss was introduced to encourage the masked image embeddings to match the unmasked image embeddings. This regularization is similar to representation invariance objectives used in self-supervised learning methods such as \cite{bardes_vicreg_2022} and \cite{chen_simple_2020}. This loss was calculated as the mean squared error between the embedding vectors of the masked and unmasked images after encoding with ResNet-18. The embedding consistency loss, denoted by $L_\text{consistency}$, is defined in Equation \ref{eq:emb}. 
\begin{equation}\label{eq:emb}
L_\text{consistency} = \frac{1}{d}
\sum_{i=1}^{d}
(z_{m,i}-z_{u,i})^2
\end{equation}
In this formulation, $d$ is the length of the embedding vector, $z_m$ is the masked embedding vector, $z_u$ is the unmasked embedding vector, and $i$ is the $i$th component of the embedding vectors.

\subsection{Joint Optimization}
The masking network and the classification network were jointly trained with the five previously described loss terms. Sparsity loss, $L_\text{sparsity}$, was needed to create the desired sparsity of masked regions. The smoothness loss, $L_\text{smoothness}$, was needed to eliminate high-frequency noise in the mask and create smoother transitions between background and structures. The consistency loss, $L_\text{consistency}$, was needed to encourage the masked image embeddings to match the unmasked image embeddings and have the same discriminative features. Finally, classification loss was needed for both masked and unmasked predictions to enable model flexibility and to allow optimal classification of both inputs. The total loss, denoted by $L_\text{total}$, was then computed from the sum of $L_\text{masked}$, $L_\text{unmasked}$, $L_\text{sparsity}$, $L_\text{smoothness}$,  and $L_\text{consistency}$ (Equation \ref{eq:total}). All of the loss terms were weighted equally.

\begin{equation}\label{eq:total}
\begin{split}
L_\text{total} =\;&
\lambda_1 L_\text{masked}
+ \lambda_2 L_\text{unmasked} \\
&+ \lambda_3 L_\text{sparsity}
+ \lambda_4 L_\text{smoothness}
+ \lambda_5 L_\text{consistency}
\end{split}
\end{equation}

\section{Experiments and Results}
\label{sec:exp_res}
We evaluated our spatial attention noise masks (SANM) across five main experiments: first, we qualitatively compared them with other popular feature attribution methods through side-by-side visual comparisons; then, we evaluated them quantitatively through classification performance, faithfulness, robustness, and an ablation study. Classification performance was measured using top-1 and top-k accuracy for each of the five classification tasks, detailed below. Faithfulness was demonstrated through insertion and deletion curves, each accompanied by its respective area under the curve score, where higher scores are better for insertion and lower scores are better for deletion. Given that our masks were capable of masking out irrelevant regions and background regions, we evaluated the robustness towards natural adversarial examples and background swapping, measured by classification accuracy. Finally, we conducted an ablation study to examine the effects and behavior of each proposed loss term in our optimization.

\subsection{Visual Comparison}
We qualitatively compared SANM side-by-side with four other state-of-the-art feature attribution methods. These methods are described below:
\begin{itemize}
    \item \textbf{GradCAM}: This method analyzes gradients of the target class with respect to the final convolutional layer and weights the feature maps \cite{selvaraju_grad-cam_2017}.
    \item \textbf{Guided GradCAM}: This method combines the concepts of GradCAM and guided backpropagation. Guided backpropagation only propagates positive gradients by using ReLU to set negative gradients to 0 \cite{springenberg_striving_2015}.
    \item \textbf{Integrated Gradients}: This is a method that accumulates gradients along a path between a baseline (black image) and the actual input image \cite{sundararajan_axiomatic_2017}.
    \item \textbf{Occlusion}: This is one of many perturbation methods that measures the difference in output based on occluding a kernel of pixels across the image \cite{zeiler_visualizing_2014}.
\end{itemize}
We evaluated the visual comparisons from five classification tasks across three datasets, which are shown in Table \ref{tab:datasets}. These datasets represent images at varying resolutions and in diverse contexts in the natural world and medical domains. The first dataset was CIFAR-100 \cite{krizhevsky_learning_2009}. Next, we evaluated on the ImageNet-1k \cite{deng_imagenet_2009} dataset. Finally, we evaluated on three subsets of the RadImageNet \cite{mei_radimagenet_2022} dataset, representing each of the three medical imaging modalities included in the overall dataset. These subsets were the Lung CT, Brain MRI, and Abdomen/Pelvis Ultrasound. We chose these subsets to isolate and demonstrate the SANM behavior on each medical imaging modality.

\begin{figure*}[]
\centerline{\includegraphics[width=0.8\textwidth]{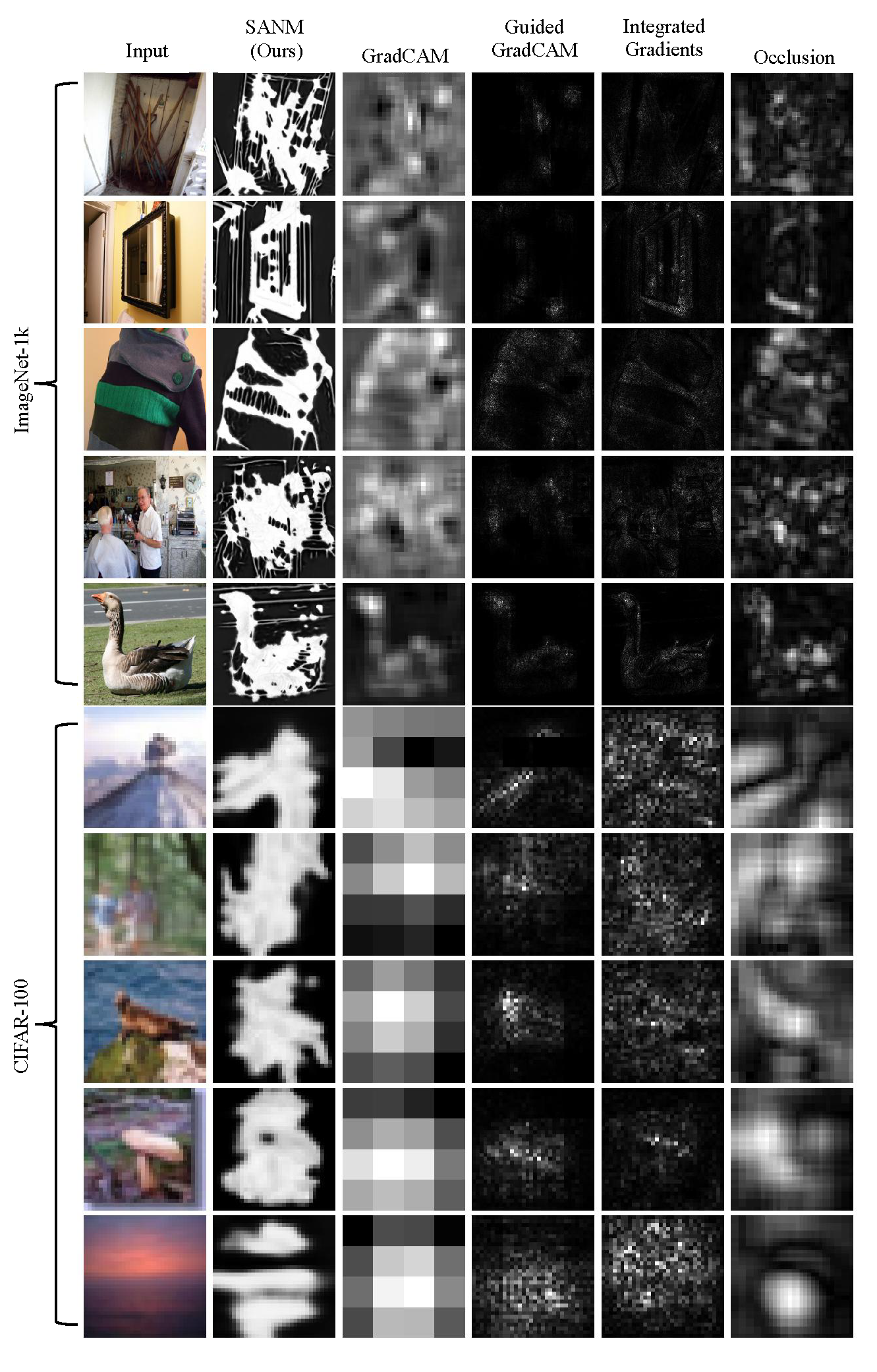}}
\caption{Example images from ImageNet-1k and CIFAR-100 with the corresponding SANM and four other feature attribution methods. Our masks tend to include the same regions highlighted by the other attribution maps, as well as additional regions. This is potentially caused by our masks being generated before classification, whereas the other attribution maps are generated post hoc and conditioned on a target class.}
\label{fig:imagenet}
\end{figure*}

Figure \ref{fig:imagenet} shows example input images from the CIFAR-100 and ImageNet-1k datasets. Our spatial attention noise masks are compared with feature attribution maps from GradCAM, Guided GradCAM, Integrated Gradients, and Occlusion. Overall, the proposed masks tend to include the same regions highlighted by the other attribution maps. In some cases, our masks include additional features not highlighted by the attribution methods. We believe this is because the masks are generated before classification, whereas the attribution methods produce class-specific explanations conditioned on the target class.

\begin{figure*}[]
\centerline{\includegraphics[width=0.8\textwidth]{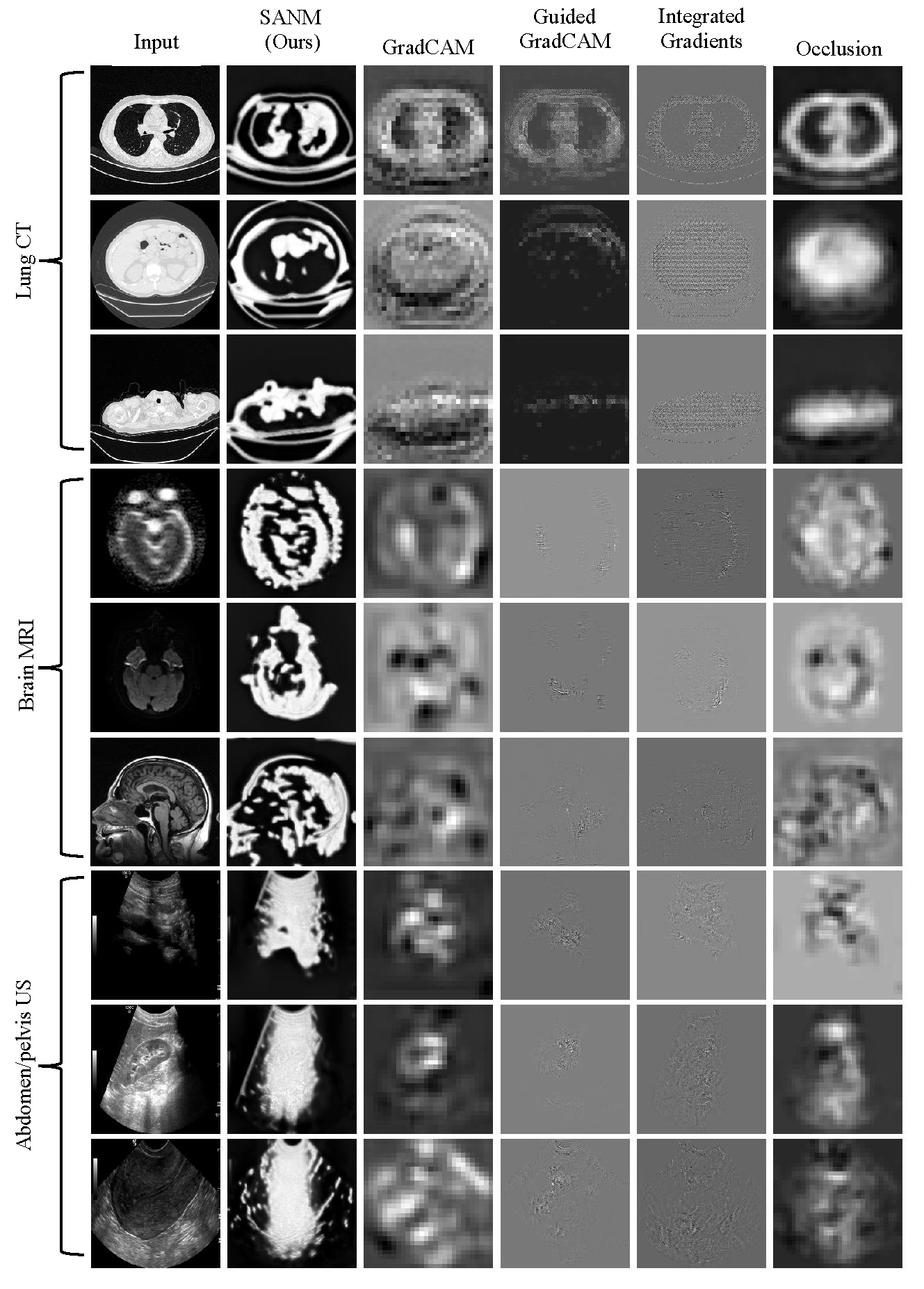}}
\caption{Example images from three medical imaging modalities in RadImageNet with the corresponding SANM and four other feature attribution methods. Our SANM provides a sharper structure than other methods but may require task-dependent sparsity tuning, since the sparsity constraint is largely satisfied by the amount of black background in many of these medical images.}
\label{fig:radimagenet}
\end{figure*}

Similarly, Figure \ref{fig:radimagenet} shows example input images from RadImageNet with our masks and the other feature attribution methods. In these medical imaging contexts, our masks provide sharper structure but are often less localized than the other maps. Again, this could be due to the class-specific explanations provided by the other methods. Interestingly, our masks exclude some regions highlighted by the other methods, particularly in the Lung CT and Brain MRI examples. For the abdomen/pelvis US task, our masks were limited by the sparsity constraint, which was satisfied by the amount of black background, often resulting in retention of the central focal point of the ultrasound scan. This shows that mask sparsity can be task-dependent and may require further tuning.

\subsection{Classification Performance}
To explore the potential trade-off between classification performance and interpretability in our method, we evaluated our SANM on the same datasets used for the visual comparisons (Table \ref{tab:datasets}). For each task, we measured the top-1 and top-k accuracies of a baseline model, as well as the masked and unmasked accuracies of our approach. We chose to use a ResNet-18 \cite{he_deep_2016} as a baseline, since it serves as the backbone classifier in our framework.

\begin{table}[]
\renewcommand{\arraystretch}{1.2}
\caption{Dataset statistics showing total images, classes, channels, and image size.}
\label{tab:datasets}
\centering
\resizebox{\columnwidth}{!}{%
\begin{tabular}{l|c|c|c|c}
Dataset & Total Images & Classes & Channels & Image Size \\ \hline
CIFAR-100 & 60k & 100 & 3 & 32x32 \\
ImageNet-1k & 1.2M & 1000 & 3 & 224x224 \\
RadImageNet-Lung CT & 152k & 6 & 1 & 224x224 \\
RadImageNet-Brain MRI & 44k & 10 & 1 & 224x224 \\
RadImageNet-Abdomen/Pelvis US & 297k & 13 & 1 & 224x224 \\
\end{tabular}
}
\end{table}

Table \ref{tab:metrics} shows the top-1 and top-k accuracies of the baseline ResNet-18 model with those of our model for masked and unmasked predictions. The accuracy of our model with unmasked images remains near baseline across all datasets, with CIFAR-100 showing the largest difference of 1.66. With masked images, our model's performance decreases, as expected, across most tasks. Overall, given that the mask significantly reduces the amount of data seen by the classifier, we still consider this an encouraging result, given the causal interpretability provided by the masks.

\begin{table}[]
\centering
\caption{Top-1 Accuracy and Top-K Accuracy for each dataset. All datasets use k=3 except ImageNet-1k uses k=5.}
\label{tab:metrics}
\resizebox{\columnwidth}{!}{
\begin{tabular}{l|l|l|l}
Dataset & Model & Top-1 Accuracy (\%) & Top-K Accuracy (\%)  \\
\hline
\multirow{3}{*}{CIFAR-100}
& Baseline  & 75.47 {(-)} & 89.70 {(-)} \\
& Ours (unmasked) & 73.81 {(-1.66)} & 88.80 {(-0.90)} \\
& Ours (masked) & 67.04 {(-8.43)} & 83.85 {(-5.85)} \\ \hline

\multirow{3}{*}{ImageNet-1k}
& Baseline  & 70.22 {(-)} & 89.62 {(-)} \\
& Ours (unmasked) & 69.76 {(-0.46)} & 89.31 {(-0.31)} \\
& Ours (masked) & 68.02 {(-2.20)} & 88.16 {(-1.46)} \\ \hline

\multirow{3}{*}{RadImageNet-Lung CT}
& Baseline  & 71.95 {(-)} & 85.34 {(-)} \\
& Ours (unmasked) & 71.20 {(-0.75)} & 84.99 {(-0.35)} \\
& Ours (masked) & 71.22 {(-0.73)} & 85.53 {(+0.19)} \\ \hline

\multirow{3}{*}{RadImageNet-Brain MRI}
& Baseline  & 88.16 {(-)} & 96.53 {(-)} \\
& Ours (unmasked) & 87.07 {(-1.09)} & 96.19 {(-0.34)} \\
& Ours (masked) & 86.02 {(-2.14)} & 95.95 {(-0.58)} \\ \hline

\multirow{3}{*}{RadImageNet-Abdomen/pelvis US}
& Baseline  & 95.80 {(-)} & 99.76 {(-)} \\
& Ours (unmasked) & 95.80 {(+0.00)} & 99.78 {(+0.02)} \\
& Ours (masked) & 95.24 {(-0.56)} & 99.71 {(-0.05)} \\ 
\end{tabular}
}
\end{table}

\subsection{Faithfulness}
We define faithfulness as the ability of an interpretation or explanation to represent the model's decision-making process. While our masks are faithful to the masked prediction by nature, we evaluated their faithfulness to unmasked predictions by computing insertion and deletion curves. Insertion and deletion curves show how the prediction changes as pixels of the input image are added or removed, in order of importance, based on the attribution map. We compared these curves with the insertion and deletion curves computed from feature attribution maps of the Occlusion method on our unmasked classifier. We used the Occlusion method as a baseline for faithfulness because it provides localized attributions and directly measures the change in output when removing a portion of the image. For the insertion curve, we measured the change in the predicted class probability as pixels were added to an image of solid noise in order of importance according to the mask and occlusion map, respectively. The noise was computed specifically for each image, as in the proposed workflow, to keep the image in distribution as much as possible. The deletion curve was calculated similarly, except pixels were removed from the input image and replaced with image-specific noise in order of importance according to the mask and occlusion map, respectively. We randomly sampled 500 images from ImageNet-1k, and the mask pixels were split into 50 equally sized bins; the same was done for the occlusion maps. Probabilities were averaged across all 500 samples.

\begin{figure}[]
\centerline{\includegraphics[width=\columnwidth]{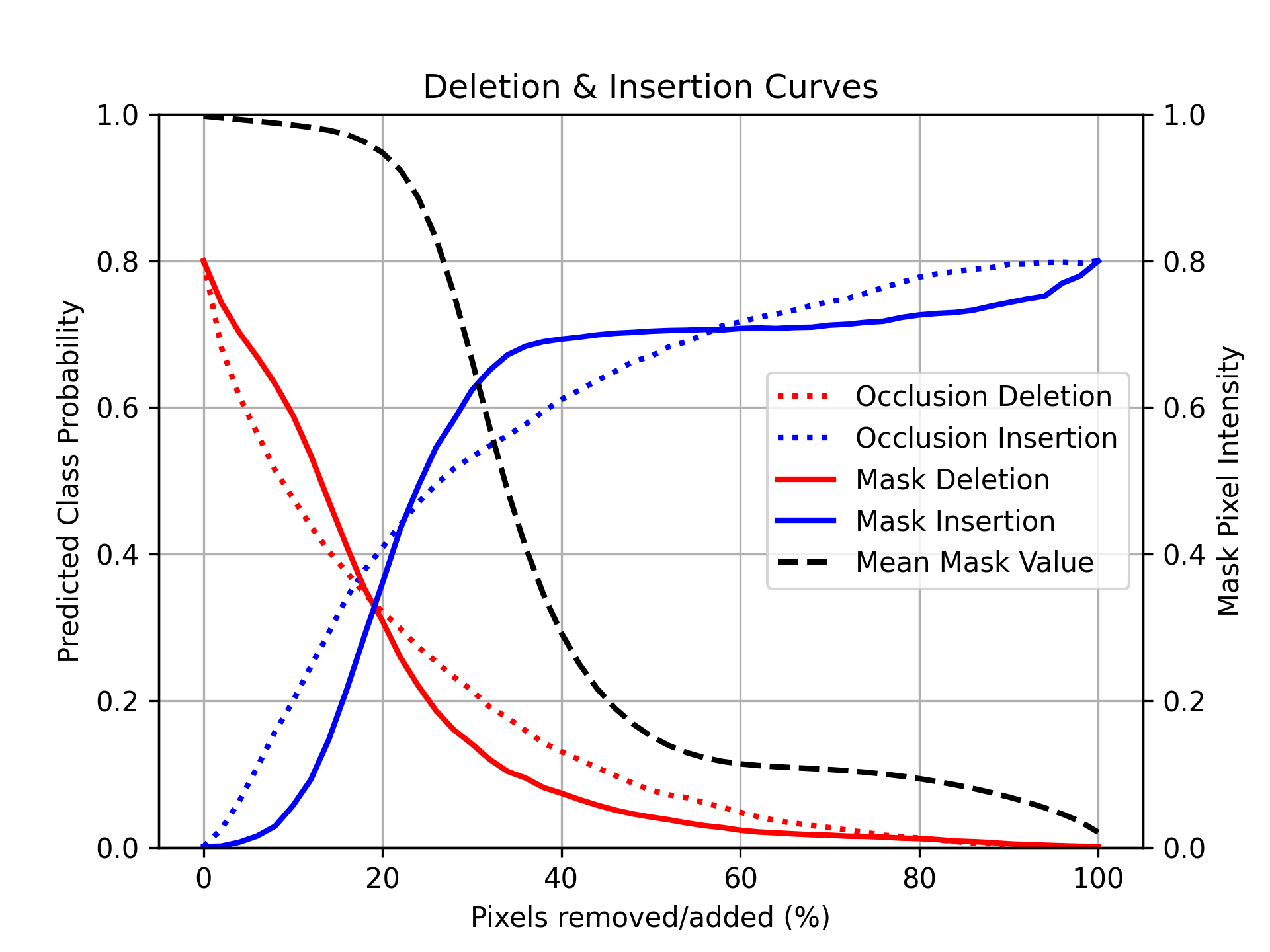}}
\caption{Average insertion (blue) and deletion (red) curves of the proposed masks (solid line) compared with the Occlusion method (dotted line) across 500 samples. The dashed black line shows the average mask pixel intensity of the inserted pixels.}
\label{fig:faith}
\end{figure}

From Figure \ref{fig:faith}, we observe that the occlusion baseline initially produces steeper insertion and deletion curves. However, our masking method surpasses it after approximately 20\% of pixels have been removed or added. For the deletion metric, both methods eventually converge to zero at a similar rate. However, in the insertion metric, the occlusion baseline overtakes the mask approach again after roughly 60\% of pixels have been restored. Notably, the mask insertion curve reaches a plateau at approximately 50\% of the pixels added, which corresponds directly to the average mask pixel intensity. This behavior indicates that the model's confidence is largely captured by the high-intensity regions selected by the mask, providing evidence that the learned masks are faithful to the prediction. Table \ref{tab:faith} further summarizes these results through the area under the insertion and deletion curves (AUC), where the proposed masks outperform the occlusion baseline on both metrics.

\begin{figure}[]
\centerline{\includegraphics[width=\columnwidth]{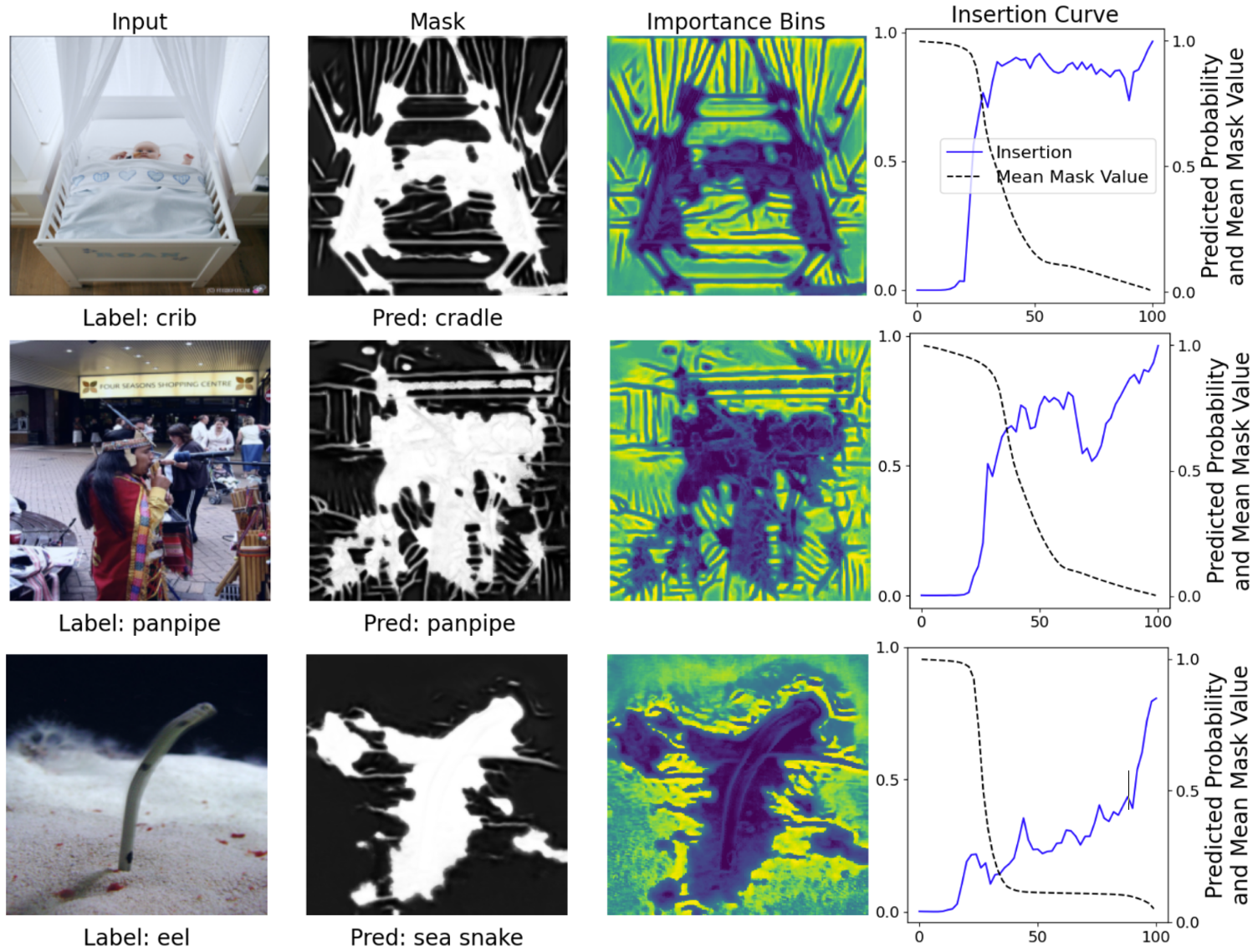}}
\caption{Selected examples demonstrating the final rise of the insertion curve shown in Figure \ref{fig:faith}. Shown are the input images with corresponding masks, insertion curves, and visualization of binned pixels in order of importance based on the mask, where the dark blue represents the most important and bright yellow represents the least important pixels.}
\label{fig:faithbins}
\end{figure}

The insertion curve of the mask in Figure \ref{fig:faith} exhibits a small increase in the final 20\% of pixels added. We illustrate why this increase may occur by providing insertion curves for individual images that exhibit this behavior, along with corresponding visualizations of the locations where the last 20\% of pixels are added. These examples are shown in Figure \ref{fig:faithbins}. We found that these final increases typically occurred in incorrect predictions, with a few exceptions. Often, the last 10 bins (or last 20\% of pixels added) created contours around the major structures in the image (see Importance Bins in Figure \ref{fig:faithbins}), so we hypothesize that the completion of the transitions between objects and background causes the final rise in the insertion curve.

\begin{table}[]
\centering
\caption{Area under the insertion curve and the deletion curve (AUC) for our masks compared to Occlusion on our model's classifier.}
\label{tab:faith}
\resizebox{\columnwidth}{!}{
\begin{tabular}{l|c|c}
Model & Insertion AUC (\%) & Deletion AUC (\%) \\
\hline

Occlusion Baseline & 57.20 & 18.66 \\
Ours & \textbf{57.46} & \textbf{16.01} \\
\end{tabular}
}
\end{table}

\subsection{Robustness}
We hypothesized that SANM would improve robustness to natural adversarial examples, where objects appear in unusual contexts, as well as to distribution shifts such as background changes. To evaluate this hypothesis, we tested our model on ImageNet-A \cite{hendrycks_natural_2021} and on a modified version of ImageNet-1K in which object backgrounds were replaced using segmentation masks from ImageNet-S \cite{gao_large-scale_2023}. The ImageNet-A dataset is a carefully filtered and curated dataset of 7,500 images that represent a subset of 200 classes from the ImageNet-1k dataset. The images are naturally occurring adversarial examples of the ImageNet-1k classes in misleading contexts, such as a dragonfly on a textured chair rather than a leaf or branch. For the background-swapped dataset, each object was composited onto a randomly selected background from a set of ten publicly available environmental scenes, including forest, jungle, lake, underwater, mountain range, tundra, desert, beach, sky, and city. 

\begin{figure}[]
\centerline{\includegraphics[width=\columnwidth]{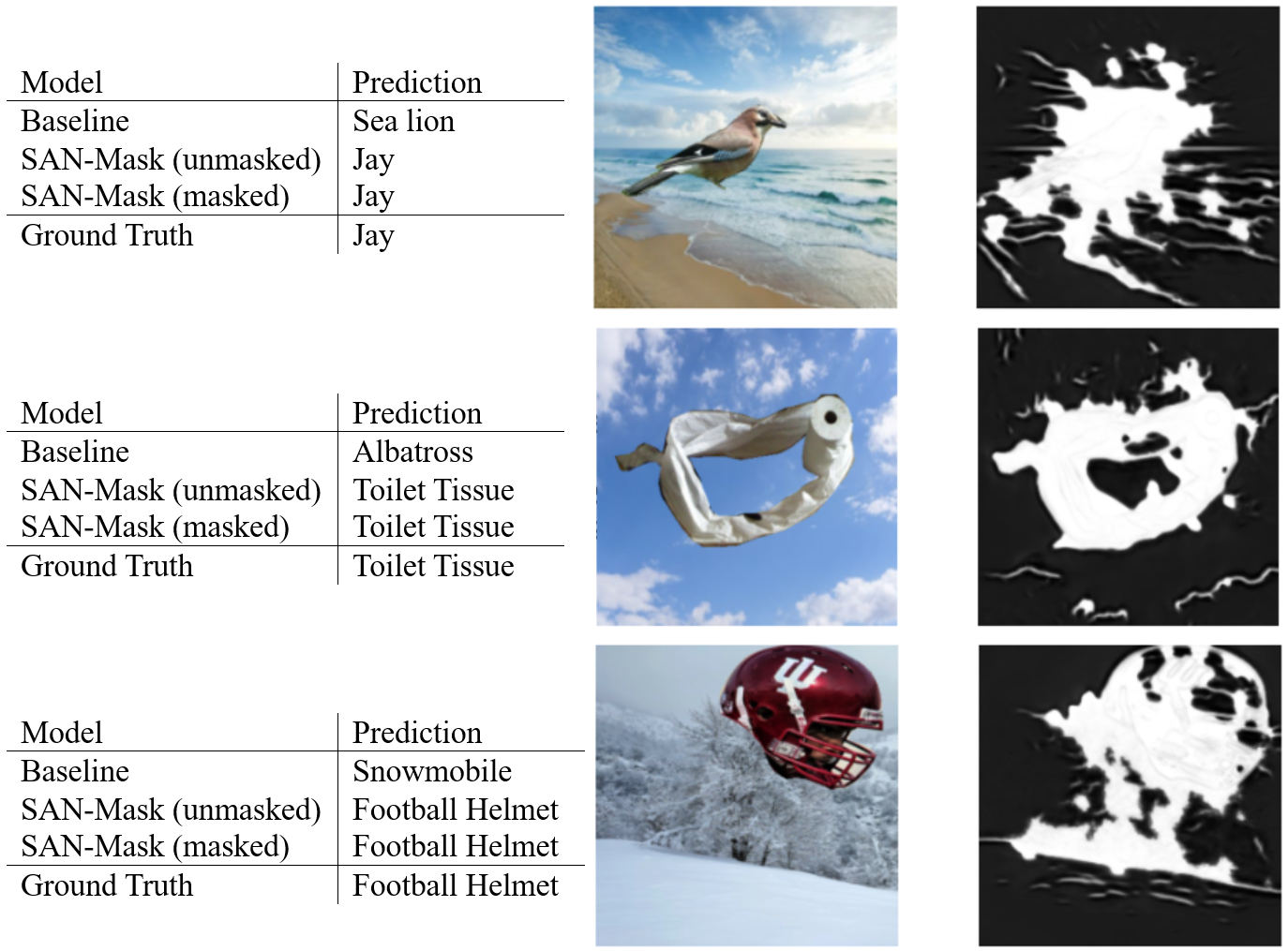}}
\caption{Selected example images and model predictions of background-swapped ImageNet-1k experiment. The baseline was often influenced by background, leading to misclassification, whereas the masked prediction maintained the correct prediction.}
\label{fig:robust}
\end{figure}

Table \ref{tab:robustness} shows the top-1 accuracies of the baseline model and our masking model on ImageNet-A and ImageNet-1k when backgrounds are swapped using the segmentations from ImageNet-S. Our model shows similar performance degradation to the baseline with unmasked images. However, it performs better on masked images, demonstrating that our masking model is more robust to distribution shifts involving background swapping. In many cases, the baseline model was strongly influenced by the background, so we selected examples that demonstrate these scenarios in which the baseline model misclassified the input and our masking model maintained the correct classification, shown in Figure \ref{fig:robust}. For example, the baseline misclassified a jaybird on a beach background as a sea lion. In contrast, our mask suppressed much of the irrelevant background information while preserving the features necessary for accurate classification.

\begin{table}[]
\centering
\caption{Top-1 Accuracy of ImageNet-A and ImageNet-1k with background swapping using segmentations from ImageNet-S.}
\label{tab:robustness}
\resizebox{\columnwidth}{!}{
\begin{tabular}{l|c|c}
Model & ImageNet-A (\%) & ImageNet-1k Background Swap (\%) \\
\hline

Baseline & 1.98 & 57.23 \\
Ours (unmasked) & 1.87 & 57.31 \\
Ours (masked) & \textbf{2.09} & \textbf{59.97} \\
\end{tabular}
}
\end{table}

\subsection{Ablation Study}
Finally, we evaluated the contribution of each loss term through an ablation study in which individual losses were removed during training, and the resulting models were assessed using CIFAR-100 classification accuracy, area under the insertion curve (AUC), and qualitative mask visualizations. These metrics were selected to demonstrate that each loss term contributes either to predictive performance, interpretability, or both.

\begin{figure}[]
\centerline{\includegraphics[width=1.0\columnwidth]{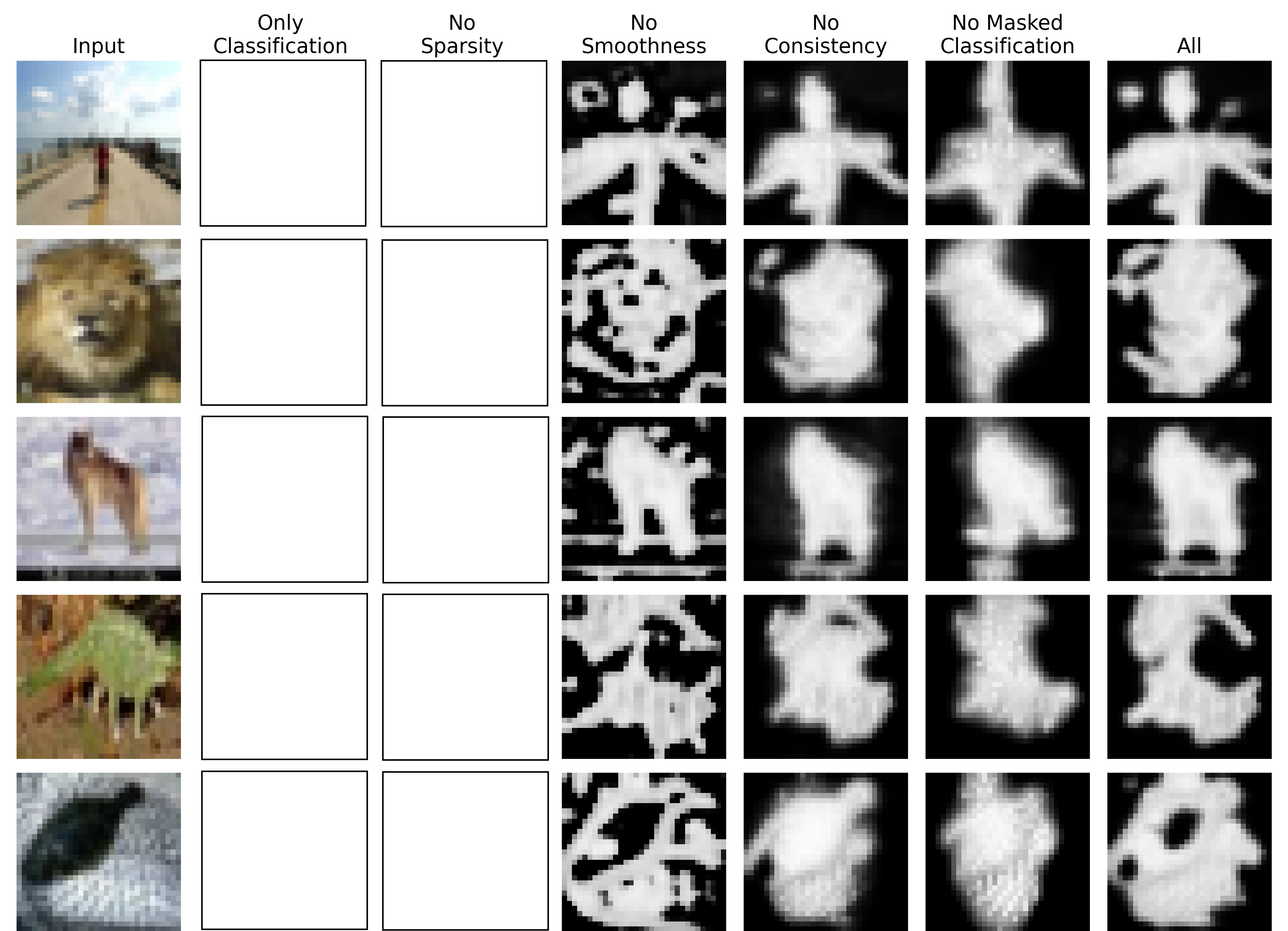}}
\caption{Resulting masks from ablation study of loss terms on CIFAR-100. Removing the sparsity loss provides no interpretability and produces uniform white masks. Smoothness loss removes high-frequency noise while maintaining structure. Consistency and masked classification losses have a smaller visual effect on masks, but improve classification performance (see Table \ref{tab:ablation}).}
\label{fig:ablation}
\end{figure}

Table \ref{tab:ablation} reports the top-1 accuracy of both masked and unmasked predictions, along with the insertion AUC, for each loss configuration. The results show that removing the sparsity loss yields the highest classification accuracy; however, interpretability largely disappears, as shown by the low insertion AUC and the nearly uniform white masks in Figure \ref{fig:ablation}. In contrast, the consistency and masked classification losses primarily influence classification performance, with minimal impact on mask structure. The smoothness loss improves both predictive performance and mask quality by removing noise and creating softer transitions. Overall, the full objective, consisting of all five loss terms, achieves the best balance between classification accuracy and interpretability, producing faithful explanations without sacrificing predictive performance.

\begin{table*}[]
\centering
\caption{Unmasked and masked classification accuracies and area under the insertion curve (AUC) for the ablation study of loss terms on CIFAR-100. Bold values represent the optimal balance of interpretability and performance for both masked and unmasked classification accuracy.}
\label{tab:ablation}
\resizebox{\textwidth}{!}{
\begin{tabular}{l|ccccc|cc|c}
Loss Terms & Sparsity & Smoothness & Consistency & Masked Classification & Unmasked Classification & Unmasked (\% Acc) & Masked (\% Acc) & Insertion AUC \\
\hline
Only Classification &  &  &  & \checkmark & \checkmark & 75.00 & 75.03 & 27.81\\
\hline
No Sparsity &  & \checkmark & \checkmark & \checkmark & \checkmark & 74.92 & 74.92 & 27.35\\
\hline
No Smoothness & \checkmark &  & \checkmark & \checkmark & \checkmark & 74.60 & 67.33 & 58.88\\
\hline
No Consistency & \checkmark & \checkmark &  & \checkmark & \checkmark & 72.98 & 65.26 & 59.16\\
\hline
No Masked Classification & \checkmark & \checkmark & \checkmark &  & \checkmark & 72.63 & 61.26 & 58.61\\
\hline
All & \checkmark & \checkmark & \checkmark & \checkmark & \checkmark & \textbf{73.81} & \textbf{67.19} & \textbf{61.06}\\
\end{tabular}
}
\end{table*}

\section{Discussion}

\subsection{Mask Behavior}
For the ImageNet-1k and CIFAR-100 datasets, our masks included the same highlighted regions as the other feature attribution methods, but often included additional regions. We believe these differences arise because our masks are generated before classification, whereas post hoc attributions are generated after prediction and are conditioned on a specific target class. This highlights an important distinction in what the explanations represent. Our masks identify features that are sufficient for the model to produce its prediction, rather than attempting to recover all features that contributed to the prediction. Consequently, the masks should not be interpreted as a complete characterization of the model's decision process. However, they provide a direct guarantee about the information available to the classifier, which can be particularly valuable in high-stakes settings where understanding the evidence supporting a prediction is critical.

Interestingly, for the RadImageNet dataset, our masks exclude some regions highlighted by the other methods, specifically in the Lung CT and Brain MRI examples. Although it is difficult to determine whether the excluded regions contribute meaningful information to the correct prediction, our approach provides a direct link between the explanation and the model's decision-making process, as the prediction can be generated solely from the masked image. Consequently, the highlighted regions represent information that was sufficient for the model to produce its prediction. As previously mentioned, our masks were limited in some of the medical contexts due to the sparsity constraint being satisfied by the large amount of black background. This suggests that the sparsity may need fine-tuning depending on the task. Additionally, we hypothesize that trivial tasks with weaker gradient signals during training may provide insufficient supervision for the mask generator to identify meaningful features. This was most evident in the Abdomen/pelvis US task, where masks were less focused than the masks generated in the more challenging classification tasks.

\subsection{Performance Trade-off and Applicability}
We acknowledge that our approach introduces a trade-off between classification performance and interpretability. Across the evaluated datasets, this trade-off was most pronounced on CIFAR-100 and RadImageNet Brain MRI, which were among the smallest datasets in terms of both image resolution and sample size, respectively. In contrast, the performance gap generally decreased across larger, higher-resolution datasets, suggesting that the impact of the masking process may become less severe as dataset size and image resolution increase. To mitigate this trade-off, we designed our framework to be parallel, thereby allowing unmasked predictions, which maintained near-baseline performance on most datasets while still enabling the generation of interpretable masks. Beyond preserving predictive performance, the parallel design enables direct comparison between the masked and unmasked predictions. The agreement between the two indicates that the masked prediction can be used with strong causal evidence that the highlighted regions are sufficient for the model's decision. Disagreement, on the other hand, suggests that information outside the masked regions contributed to the unmasked prediction and may warrant further inspection. Although some reduction in accuracy is expected, considering that the mask significantly reduces the amount of data seen by the classifier, the observed performance remained competitive across most tasks. These results are encouraging, particularly in applications where the ability to identify the specific evidence supporting a prediction is an important requirement alongside predictive performance.

\section{Conclusion}
\label{sec:conlusion}
In this work, we introduced a spatial attention noise mask framework for causal interpretability. Unlike traditional interpretability approaches that rely on passive post hoc analysis, our method actively learns interpretable explanations by jointly training a mask generator and classifier with a novel regularization scheme that replaces masked regions with noise prior to classification. Furthermore, while many causal interpretability methods focus on counterfactual explanations that identify which features were unnecessary for a prediction, our approach provides intuitive causal explanations by highlighting the features that were sufficient to support a model's decision.

Experimental results demonstrate that the proposed spatial attention noise masks yield explanations competitive with state-of-the-art feature attribution methods while maintaining strong predictive performance. Despite substantially reducing the information available to the classifier, masked predictions showed only a modest decrease in accuracy, while unmasked predictions remained near baseline performance. We further showed that the learned masks are more faithful to the model's predictive behavior than an occlusion-based baseline and provide an additional robustness benefit by reducing sensitivity to background changes.

A limitation of the proposed approach is a reduction in classification performance, making it most applicable in domains where interpretability is critical. However, the parallel masked and unmasked design mitigates this trade-off by providing strongly interpretable masked predictions alongside near-baseline unmasked performance. Moreover, the learned masks can be used as a general interpretability tool in the same manner as conventional feature attribution methods. Additionally, our approach is limited by our regularization priors, which target object detection tasks and classification tasks where the class-specific features are distinguishable objects, rather than classification tasks where discriminative features may be texture-based or visually diffuse patterns rather than distinct objects. Future work will explore alternative mask-generation architectures, classifier backbones, and evaluations across a broader range of datasets and tasks.

\section*{Acknowledgments}
This work was supported in part by NSF Project (1934966, 2029260, 2027208, 2330891) and NIH Project P20 GM121342.

\bibliographystyle{IEEEtran}
\bibliography{refs}

\end{document}